\documentclass{article}
\usepackage{spconf,amsmath,graphicx}

\title{OPTICAL FLOW BASED BACKGROUND SUBTRACTION WITH A MOVING CAMERA: APPLICATION TO AUTONOMOUS DRIVING}
\name{Sotirios Diamantas and Kostas Alexis\thanks{This material is based upon work supported by the "Intelligent Mobility: Living Labs" project funded by the Nevada Governor's Office for Economic Development.}}
\address{Autonomous Robots Lab\\
	Department of \\Computer Science \& Engineering\\
	University of Nevada, Reno\\
	1664 N. Virginia St, Reno, NV, 89557, USA\\
	Email: \{sdiamantas, kalexis\}@unr.edu}
\begin{document}
%
\maketitle
\begin{abstract}
In this research we present a novel algorithm for background subtraction using a moving camera. Our algorithm is based purely on visual information obtained from a camera mounted on an electric bus, operating in downtown Reno which automatically detects moving objects of interest with the view to provide a fully autonomous vehicle. In our approach we exploit the optical flow vectors generated by the motion of the camera while keeping parameter assumptions a minimum. At first, we estimate the Focus of Expansion, which is used to model and simulate 3D points given the intrinsic parameters of the camera, and perform multiple linear regression to estimate the regression equation parameters and implement on the real data set of every frame to identify moving objects. We validated our algorithm using data taken from a common bus route.
\end{abstract}

\section{Introduction}
\label{sec:intro}

Background subtraction is a fundamental problem in the field of computer vision and it has a number of applications relating to object detection and tracking, object segmentation and classification, change detection identification, among others. Background subtraction relates to the identification of objects that are moving in a scene while the images perceived are obtained from a static camera. Although significant contributions have enabled background subtraction algorithms to efficiently deal with the problem of background subtraction using a fixed, non-moving camera, little has been done to tackle the problem of background subtraction using a moving camera. In this paper, we propose a new algorithm for moving object detection using a single moving camera without the need to use supplementary sensors such as Inertial Measurement Units (IMU), stereo vision, or depth sensors.

\begin{figure}[t]
	\centerline{\includegraphics[width=1\columnwidth]{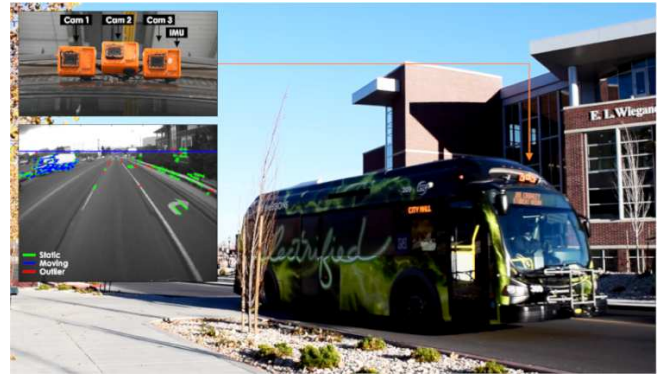}}
	\caption{Electric bus with camera sensors and IMU onboard}
	\label{bus}
\end{figure}

The algorithm we present in this paper has been tested in real-world environment and the results obtained show its robustness in terms of accuracy in spite of the high speed camera motion due to the motion of the vehicle. A multi-camera rig consisting of three cameras and an IMU has been mounted on an electric bus and images are obtained in average traffic conditions in the area between campus and downtown Reno. The employed vehicle platform, namely a Proterra EcoRide BE35 electric bus, is depicted alongside the developed camera rig. The data from one of these cameras are employed to verify the proposed method for optical flow-based background subtraction with a moving camera. Figure \ref{bus} provides a pictorial representation of the electric bus along with the sensors onboard.
Traditional methods model pixel intensities over time and detect when a scene changes over time. On the other hand, background subtraction from a moving camera is a problem that has risen during the past few years especially due to the emergence of autonomous vehicles whose guidance requires the detection of moving objects in their field of view to avoid collisions and plan a safe path.

The paper consists of five sections. In the next section, a background literature review is presented relating to research on background subtraction. In Section 3, we present the methodology followed and the various steps involved in our proposed algorithm. In Section 4, the results from this research are presented along with real world examples. Finally, in the last part of this paper, a conclusions section is presented along with a discussion on the results obtained.

\section{BACKGROUND WORK}

In contrast to the traditional methods for background subtraction which employ Gaussian Mixture Models (GMMs) \cite{KaewTraKulPong2002, DBLP:conf/icpr/Zivkovic04, ZivkovicHeijden2006} these usually apply to cases where the camera is static. For moving cameras different methodologies, however, are adopted. Following are examples of implementations that have been carried out using moving cameras. In \cite{Elqursh2012} an online method is presented that makes use of long term trajectories along with a Bayesian filtering framework to segment background from the foreground in each frame. In \cite{Sheikh2009} the authors exploit the fact that all trajectories to static areas in a scene lie in a three dimensional subspace and is used to differentiate between foreground and background objects. No 3D reconstruction of the scene is required but rather a sparse model of the background is created from feature trajectories. In \cite{Zhu2017} the problem of moving camera background subtraction is not approached as binary problem but rather as a multi-label segmentation problem by modeling different foreground objects in different layers. In addition, Bayesian filtering is employed to infer a probability map and a multi-label graph-cut based on Markov Random Field is used for labeling. Finally, in \cite{Li2015} a saliency based method is presented using the SIFT flow field of moving objects.


\section{METHODOLOGY}

In this section a description is given about the methods followed to tackle the problem of background subtraction with a moving camera. At first, the camera rig with the sensors is mounted on an electric bus manufactured by Proterra (EcoRide BE35) and operated by the Regional Transportation Commission (RTC) which is employed in the city of Reno. The data was collected during the normal operation hours of the bus following the standard route. The cameras used are \emph{FLIR PointGrey Chameleon3} (sensor: ON Semi PYTHON 1300 CMOS, 1/2'') with a resolution of $1280\times1024$ set at $20$ fps. Two cameras are gray scale and another one is a color one. The IMU unit is an $um7$ which is not used for the purpose of this research. For recording camera data we have used the Robot Operating System (ROS). 

\begin{figure}[t]
	\centerline{\includegraphics[width=3.0in]{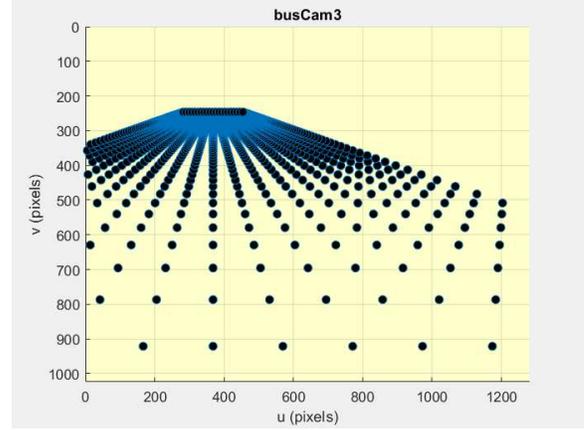}}
	\caption{Projection of 3D points based on camera intrinsic parameters and Focus of Expansion for each frame. FOE at current Figure is FOE=(368,216)}
	\label{project}
\end{figure}

First, the camera used for the background subtraction problem was calibrated and the intrinsic parameters were obtained. Optical flow vectors from the camera motion produce the so-called Focus of Expansion (FOE). FOE is calculated using the Lucas-Kanade (LK) algorithm \cite{LucasKanade1981}. Based on the intrinsic parameters of the camera, a series of 3D points are simulated and projected onto a simulated image plane like the one used by the camera sensor. The camera height from the ground plane is fixed and thus is used to simulate as accurately as possible moving objects that lie on the ground plane, i.e., the road. In total, 2000 3D points have been simulated. Figure \ref{project} shows a pictorial representation of the projected 3D points with $FOE=(368,216)$

The LK algorithm is applied to two time adjacent image frames and the \emph{leave-one-out} resampling method is used to estimate the FOE. Subsequently, a mathematical description of the \textit{leave-one-out} method used to estimate the FOE point from optical flow vectors is presented.

The convergence of optical flow vectors at point $P$ occurs when a system of linear equations is minimized. Vectors, therefore, serve as linear equations. Equations (\ref{leastSquaresMethod}) and (\ref{leastSquaresMethod_II}) present an example of two vectors,
\begin{equation}
\label{leastSquaresMethod} P\in \Omega_1=\{h\in \Re^2| \underbrace{(v_1-r_1)^T}_{\alpha_1}h=\underbrace{v_1^T\cdot r_1-||r_1||^2\}}_{\beta_1}
\end{equation}
\begin{equation}
\label{leastSquaresMethod_II} P\in \Omega_2=\{h\in \Re^2| \underbrace{(v_2-r_2)^T}_{\alpha_2}h=\underbrace{v_2^T\cdot r_2-||r_2||^2\}}_{\beta_2}
\end{equation}
where $r$ is the position of the vector, and $v$ is a point on a line that is perpendicular to the optical flow vector. The following equations, (\ref{leastSquaresMethod_V}) and (\ref{leastSquaresMethod_VI}), show the process for $n-1$ optical flow vectors. Noise in the system is denoted by $\epsilon_i$.
\begin{equation}
\begin{array}{l}
\displaystyle h\alpha_1 + \epsilon_1=\beta_1 \\
\displaystyle h\alpha_2 + \epsilon_2=\beta_2 \\
\displaystyle \hspace{10mm} \vdots \\
\displaystyle h\alpha_{n-1} + \epsilon_{n-1}=\beta_{n-1}
\end{array}
\label{leastSquaresMethod_V}
\end{equation}
\begin{equation}
\label{leastSquaresMethod_VI} h\in argmin \sum_{i=1}^{n-1}(h\alpha_i-\beta_i+\epsilon_i)^2
\end{equation}
\begin{equation}
\label{leastSquaresMethod_III} \underbrace{\left(\sum_{i=1}^{n-1}\alpha_i \alpha_i^T+\epsilon_i\right)}_{C}h=\underbrace{\left(\sum_{i=1}^{n-1}\alpha_i \beta_i\right)}_{\gamma}
\end{equation}
\begin{equation}
\label{leastSquaresMethod_IV} h=C^{-1}\gamma
\end{equation}
The \emph{leave-one-out} sampling works by taking out one optical flow vector from the sample while estimating point $P$ using the remaining $n-1$. The Euclidean distance, $d(p_j,P_i)$, is then calculated between point $p_j$ of the removed vector and the convergence point, $P_i$. The process is repeated $n$ times, which is the number of optical flow vectors. This results in a set of $n$ distances. The optical flow vectors that fall above a certain threshold, in this case, beyond the 90th percentile, are considered as outliers. Using the inliers and the \emph{leave-one-out}, eqns. (\ref{leastSquaresMethod_V})-(\ref{leastSquaresMethod_IV}) are used to find the convergence point. The convergence point, $P$, is thus, denoted by $h$, in Eqn.(\ref{leastSquaresMethod_IV})

The purpose of estimating the FOE and camera intrinsic parameters is to as accurately as possible model and simulate the set of 3D points. The set of 3D points are projected onto the image plane and an optical flow pattern is obtained using a constant-velocity camera model. The optical flow vector magnitudes at twice the velocity, i.e., $100$ km/h (from $50$ km/h) produce vectors at twice the magnitude, thus a linear relationship can be realized. A multiple linear regression is performed between two explanatory variables ($x_1$ and $x_2$; which are pixel values along x- and y-axis) and a response variable (magnitude of optical flow vectors) which is used to fit a linear equation to observed data (Eqn. \ref{multipleLE}); in our case the magnitude of optical flow vectors between two consecutive time stamps is given with a camera velocity of $50$ km/h. Uniform sampling is performed in the optical flow vector data set from the camera with the view to estimate the linear fit model. Given the linear relationship between the different camera velocities we can even estimate the speed of the vehicle by using the ratio between the predicted linear model and the observed one derived from the uniform sample space.

\begin{equation}
\label{multipleLE} y=\beta + \beta_1x_1 + \beta_2x_2
\end{equation}

\section{RESULTS}

\begin{figure*}
	\begin{center}
		\begin{tabular}{cccc}
\includegraphics[width=1 in, height=0.8 in]{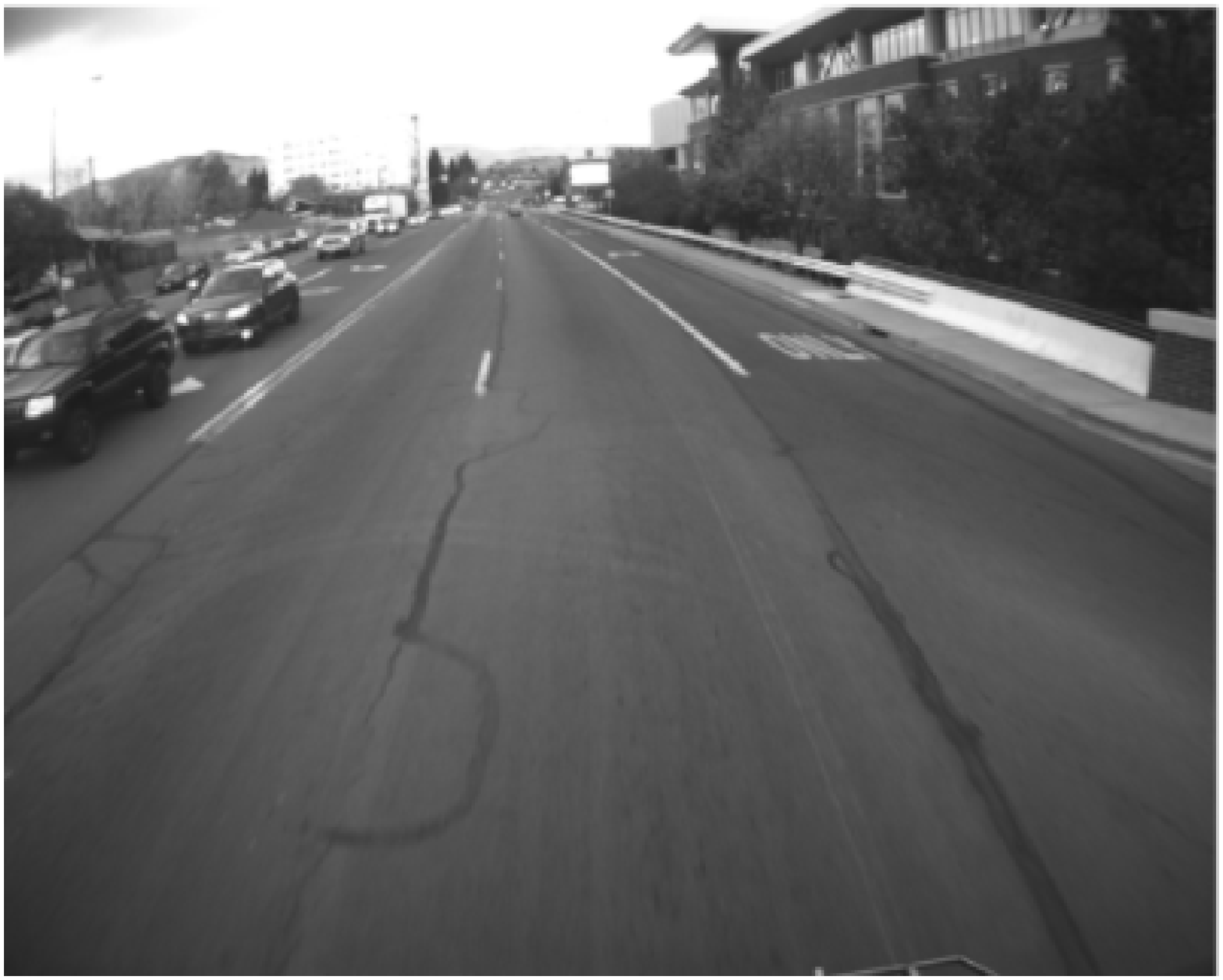}&\includegraphics[width=1 in, height=0.8 in]{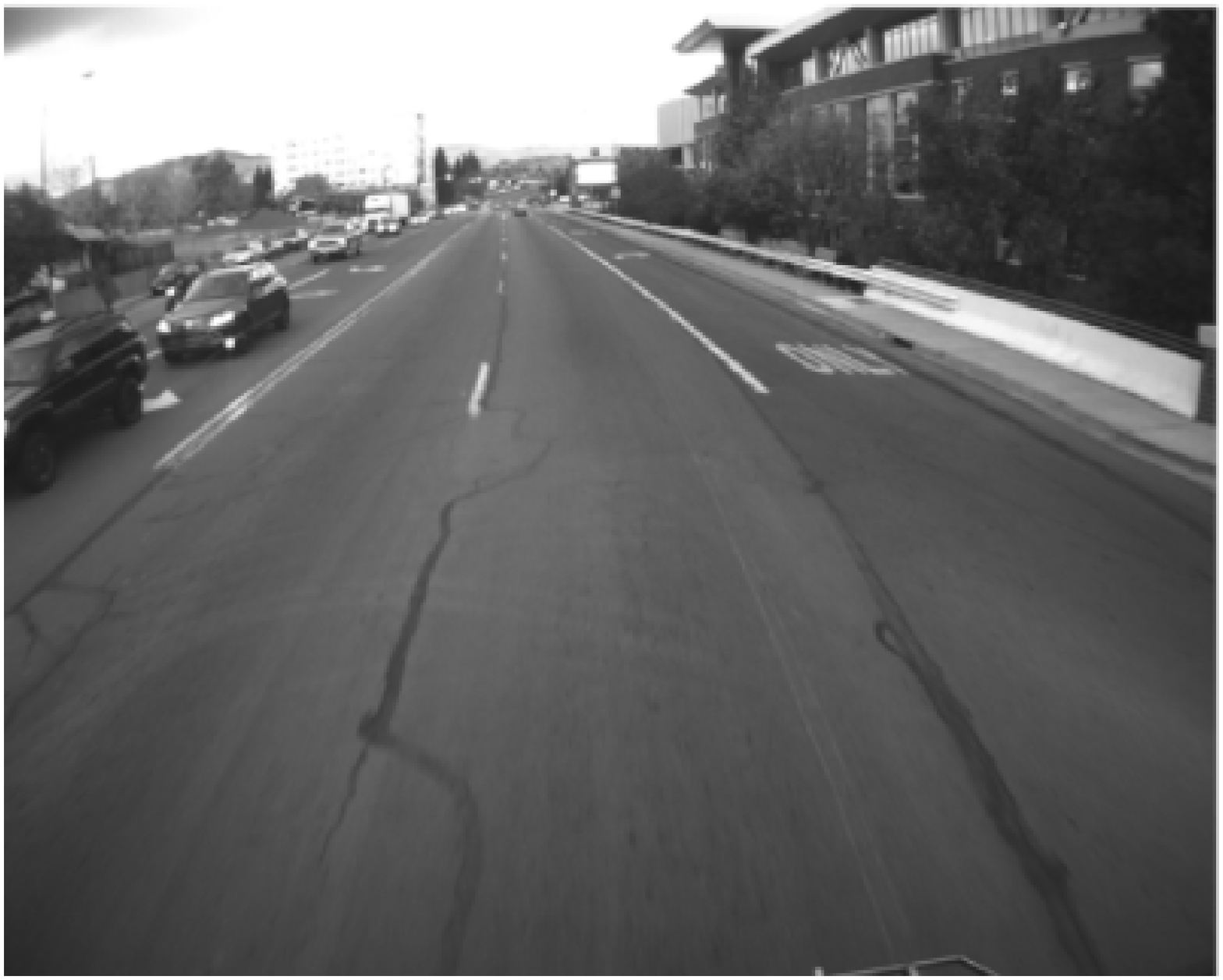}&\includegraphics[width=1 in, height=0.8 in]{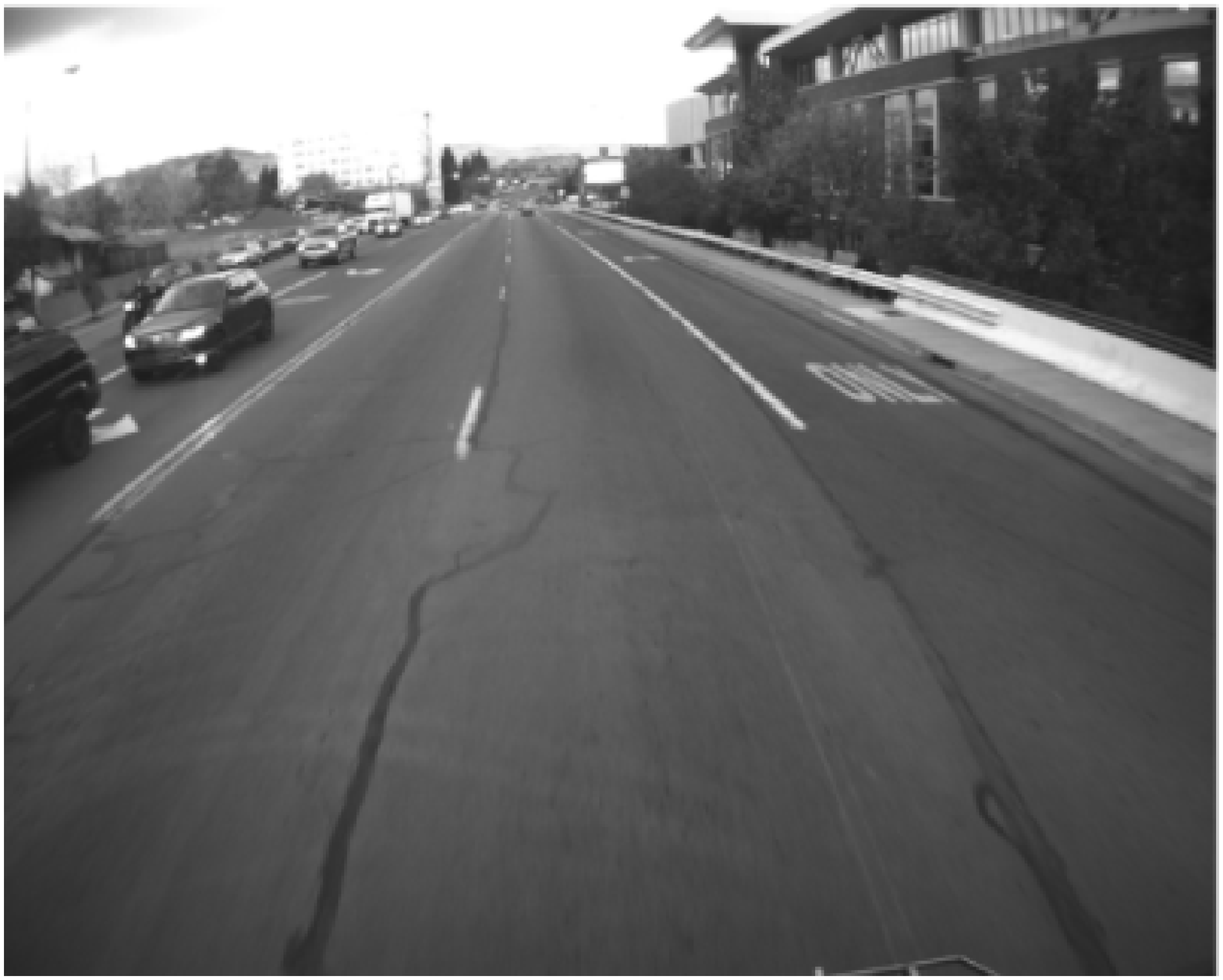}&\includegraphics[width=1 in, height=0.8 in]{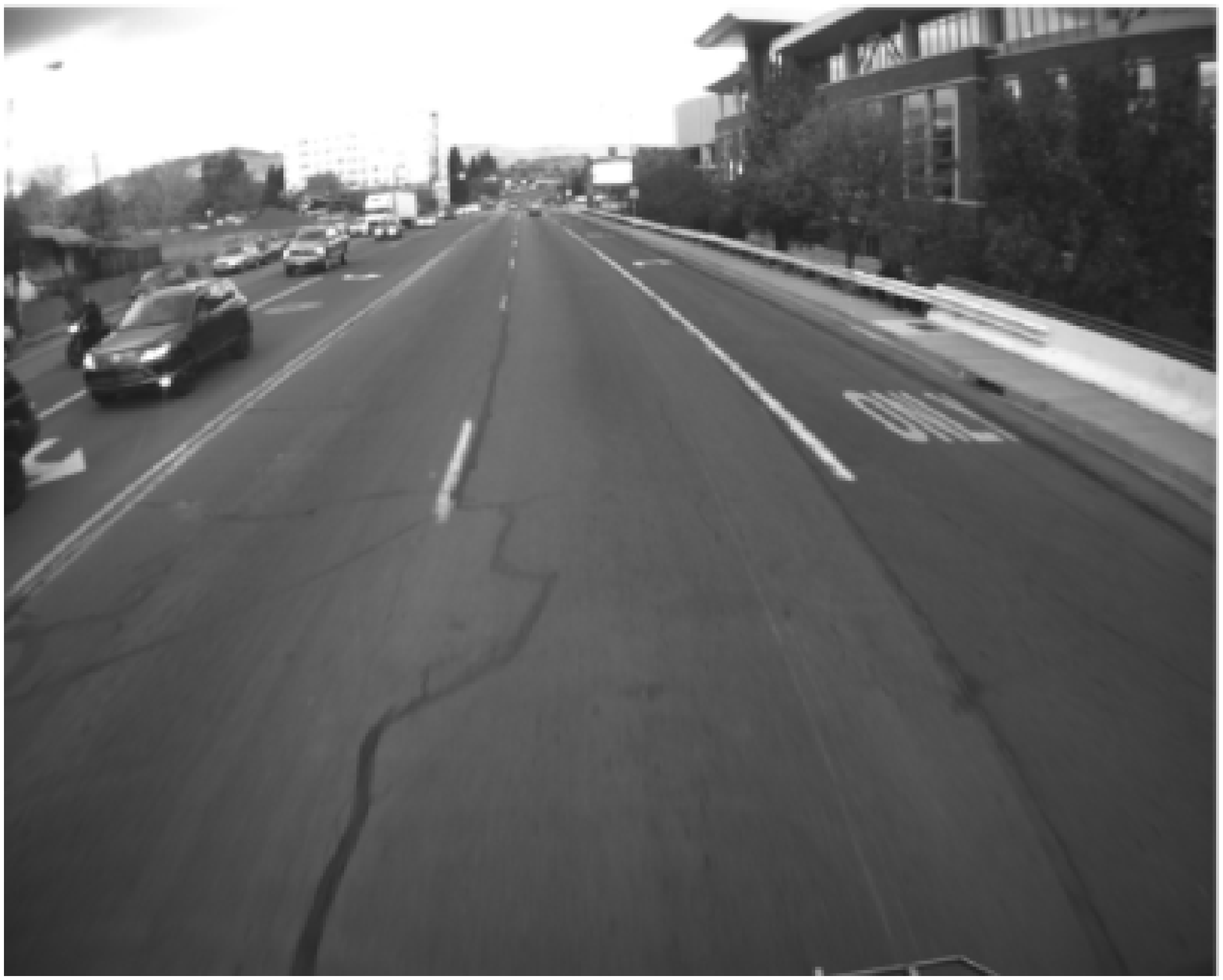}\\
			{\small (a)}&{\small (b)}&{\small (c)}&{\small (d)}\\
\includegraphics[width=1.4 in, height=1 in]{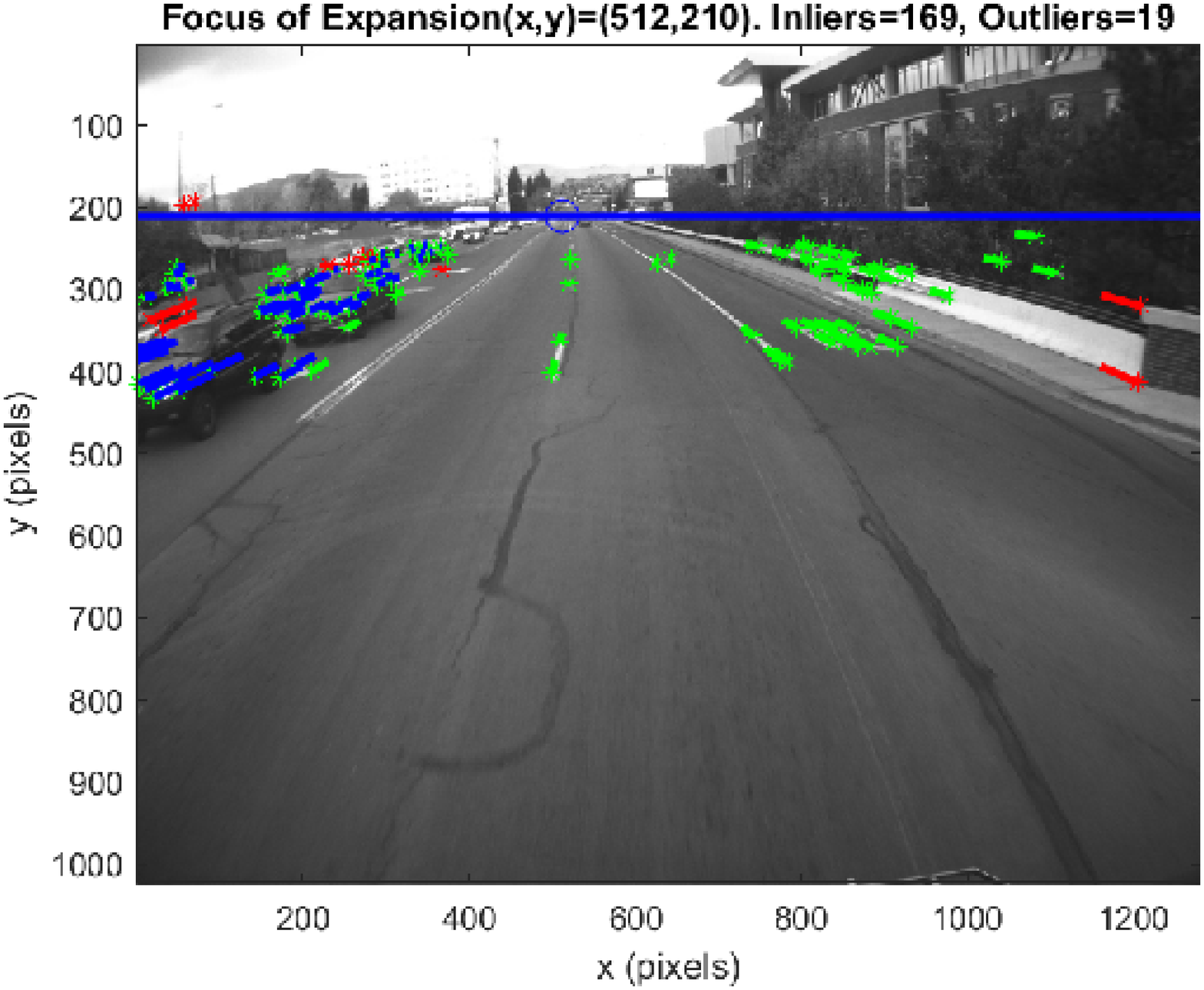}&\includegraphics[width=1.4 in, height=1 in]{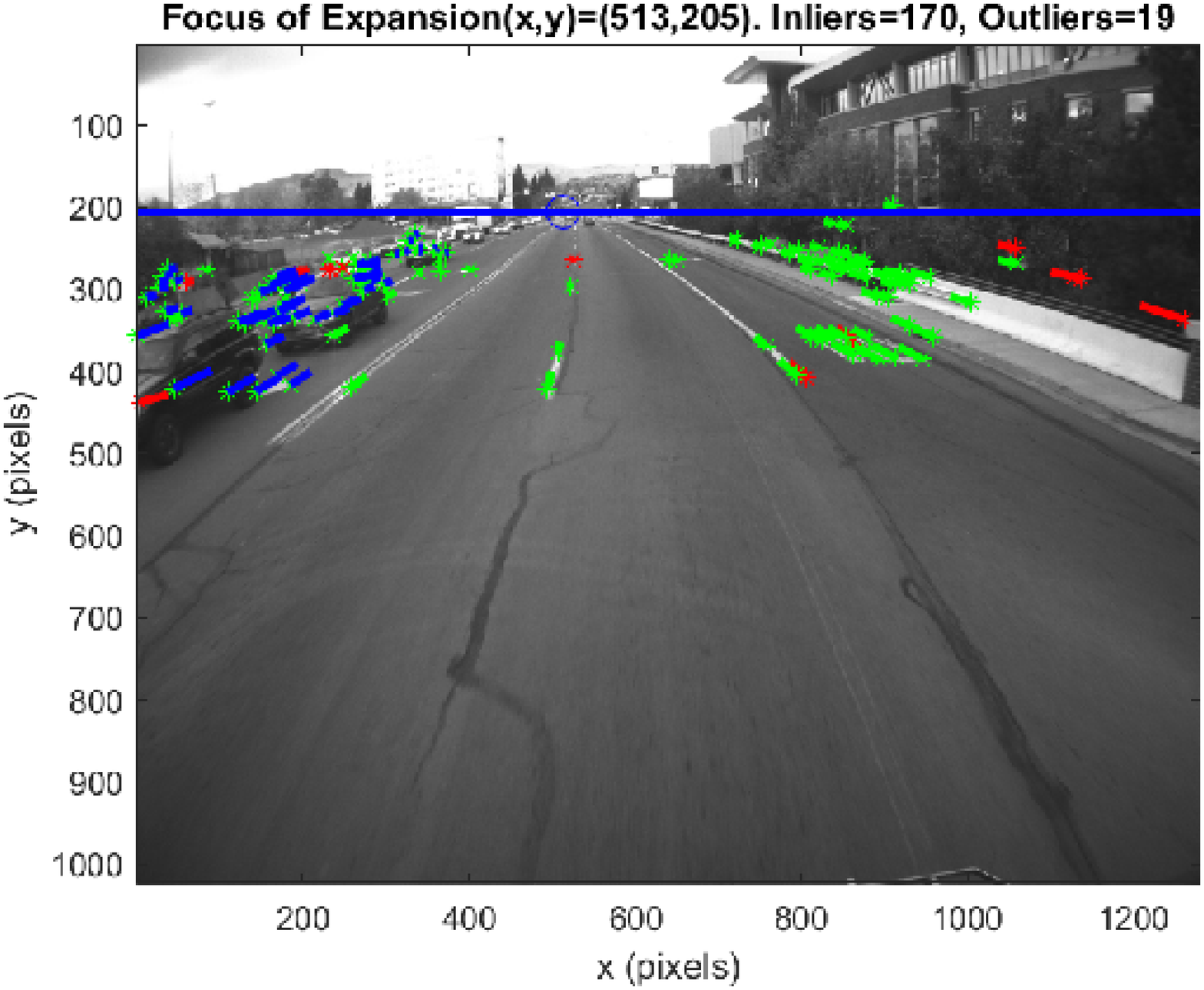}&\includegraphics[width=1.4 in, height=1 in]{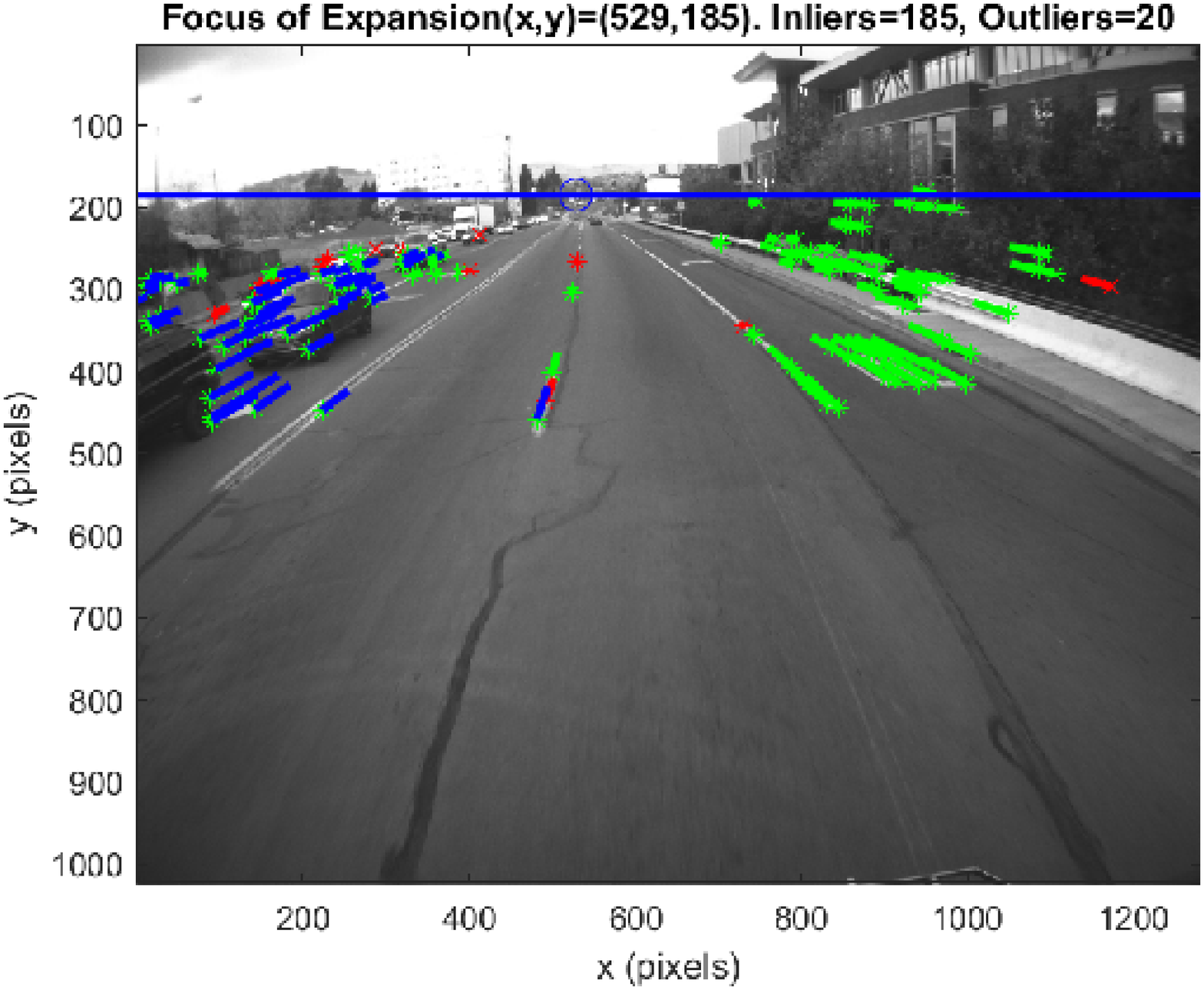}&\includegraphics[width=1.4 in, height=1 in]{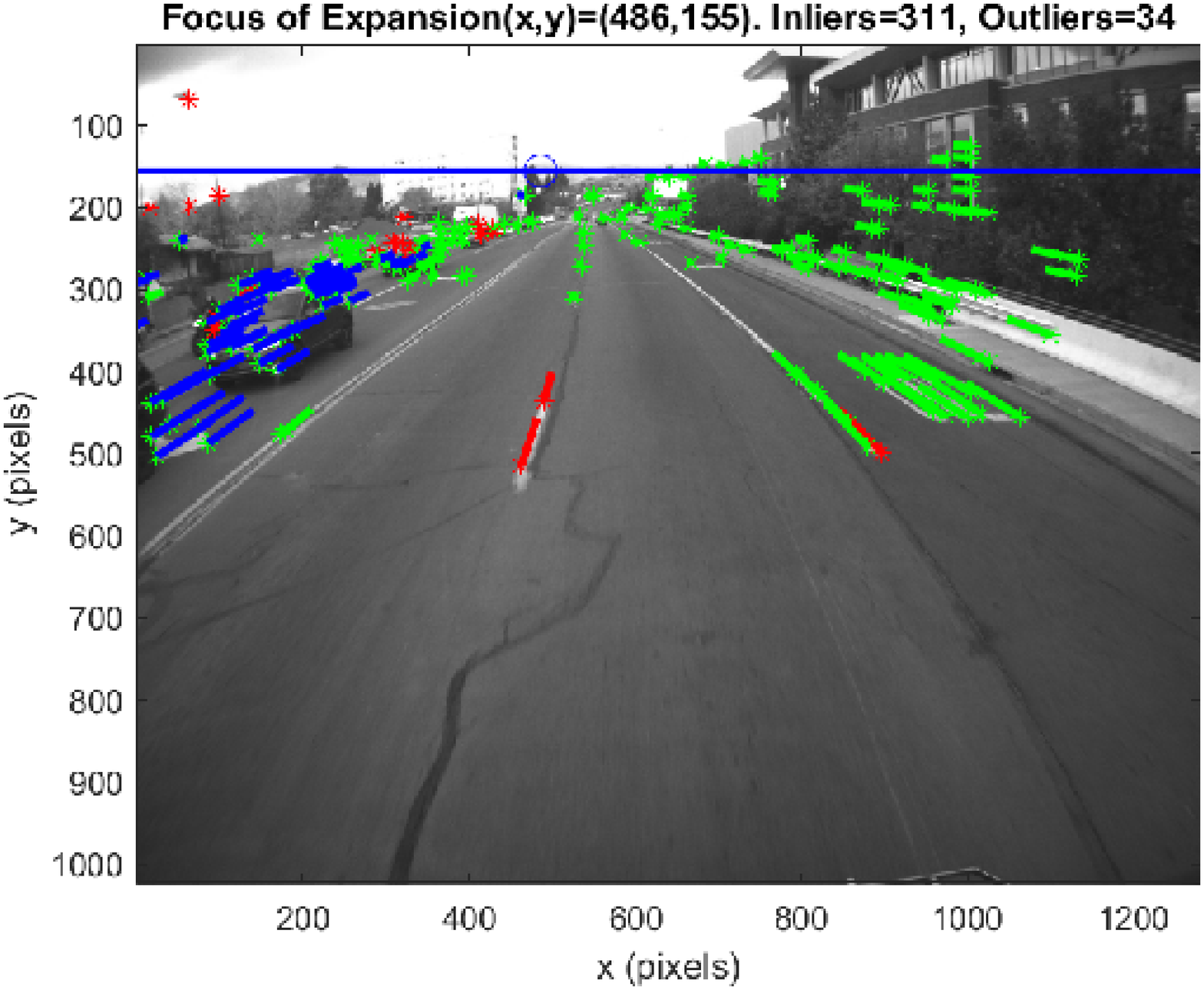}\\
			{\small (e)}&{\small (f)}&{\small (g)}&{\small (h)}\\
		\end{tabular}
	\end{center}
	\caption{(a) - (d) Raw images; (e) - (h) Results from the proposed background subtraction method. Red vectors denote outliers; Green vectors denote inliers; Blue vectors denote moving objects.}
	\label{results_one}
\end{figure*}

\begin{figure*}
	\begin{center}
		\begin{tabular}{cccc}
\includegraphics[width=1 in, height=0.8 in]{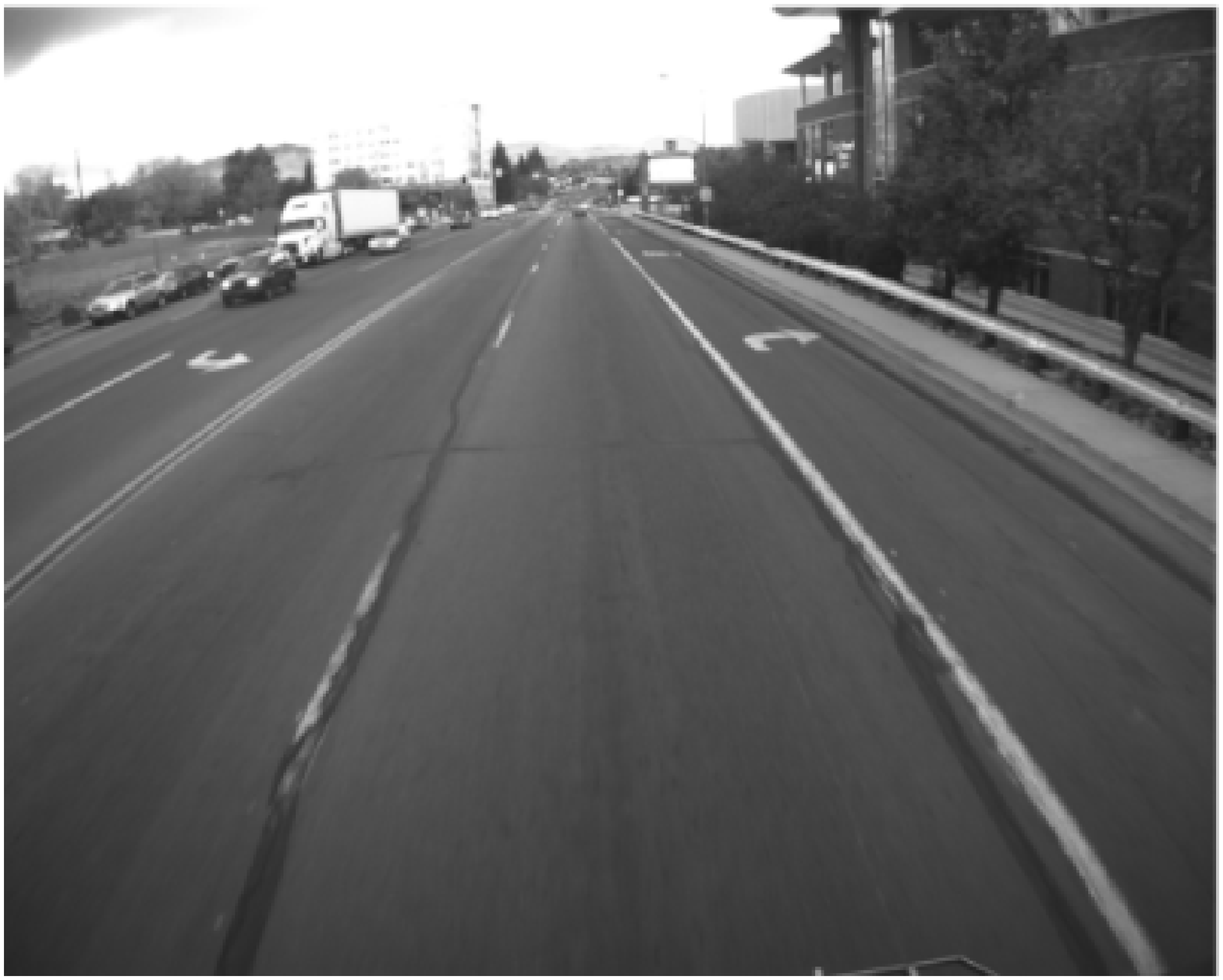}&\includegraphics[width=1 in, height=0.8 in]{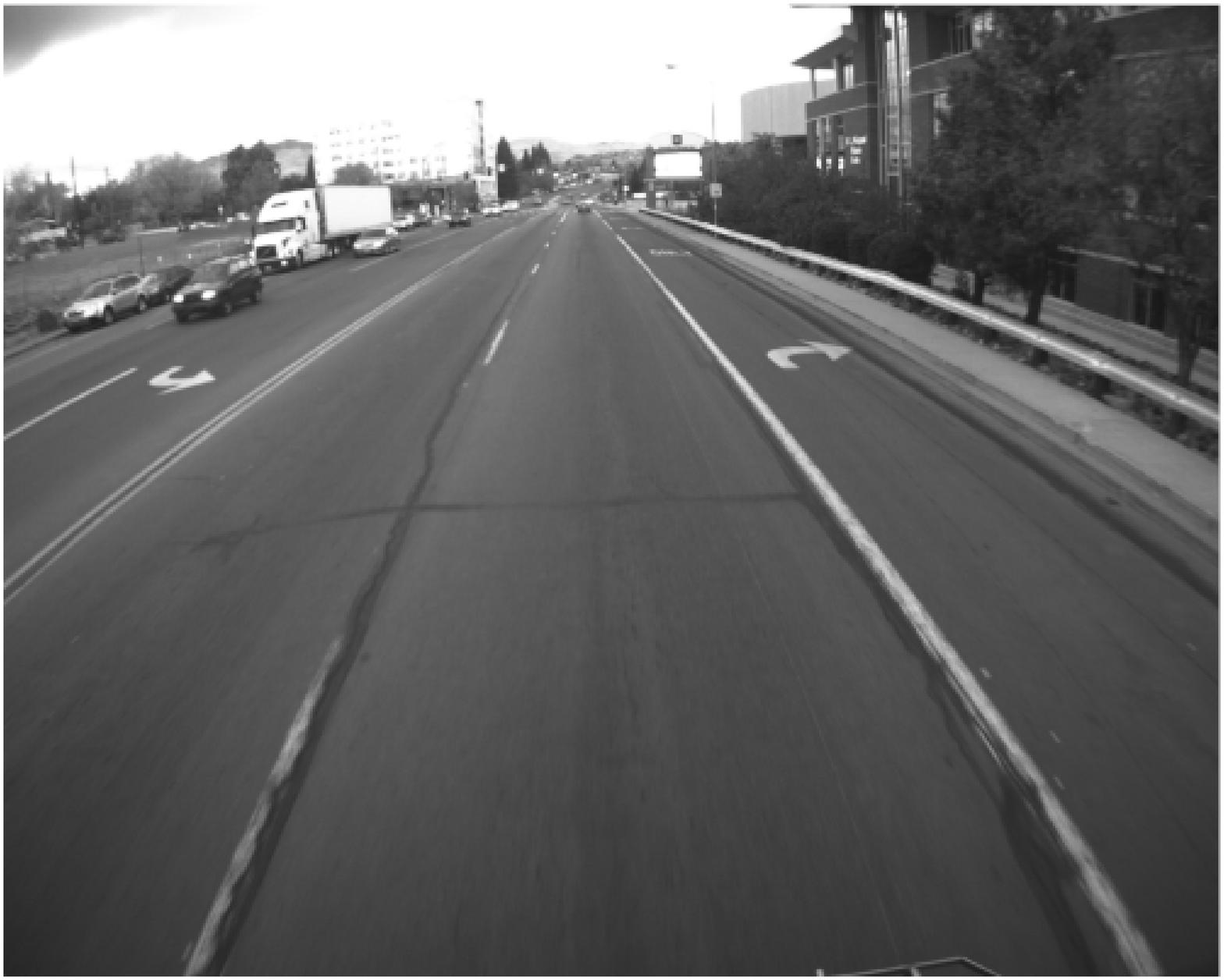}&\includegraphics[width=1 in, height=0.8 in]{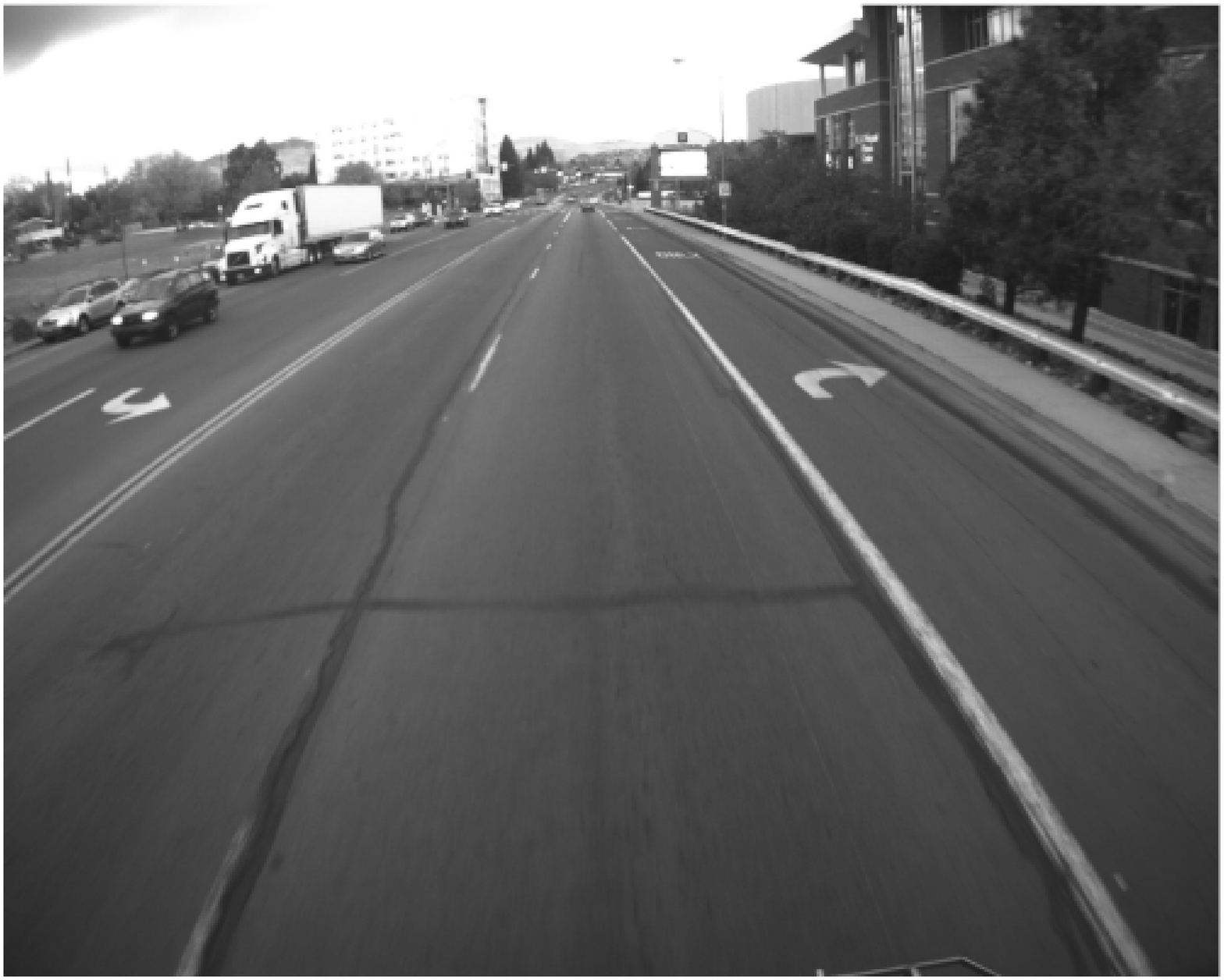}&\includegraphics[width=1 in, height=0.8 in]{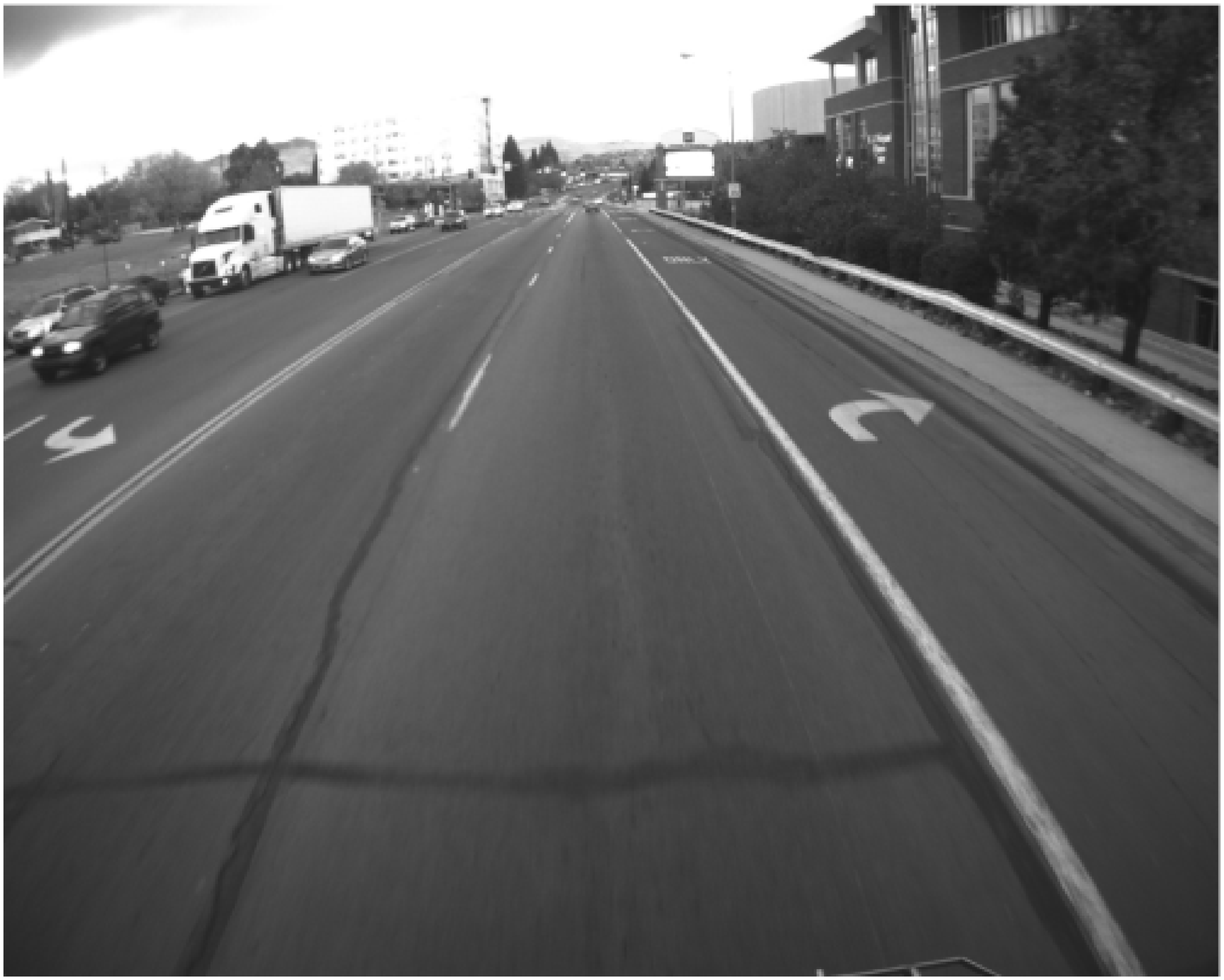}\\
{\small (a)}&{\small (b)}&{\small (c)}&{\small (d)}\\
\includegraphics[width=1.4 in, height=1 in]{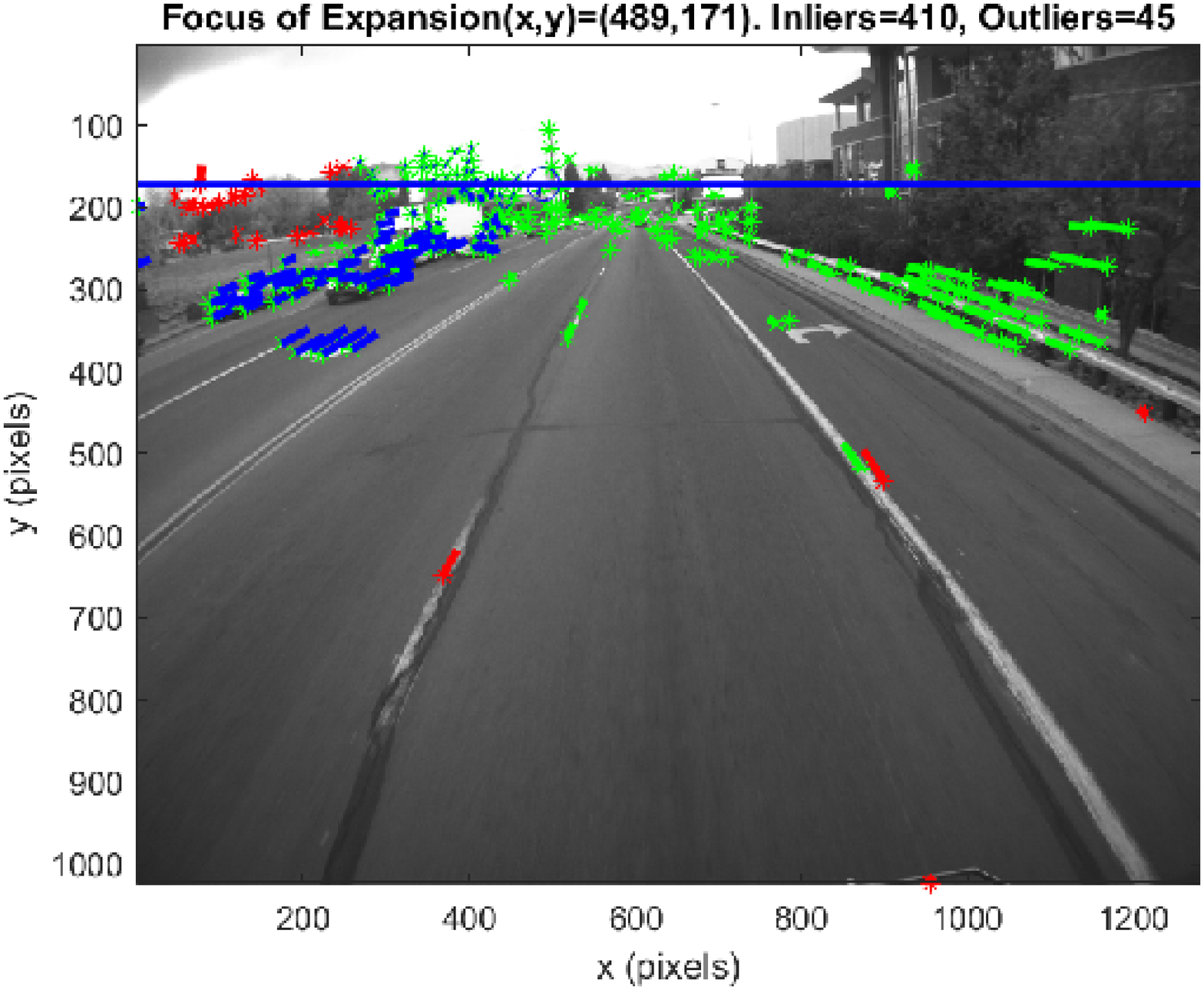}&\includegraphics[width=1.4 in, height=1 in]{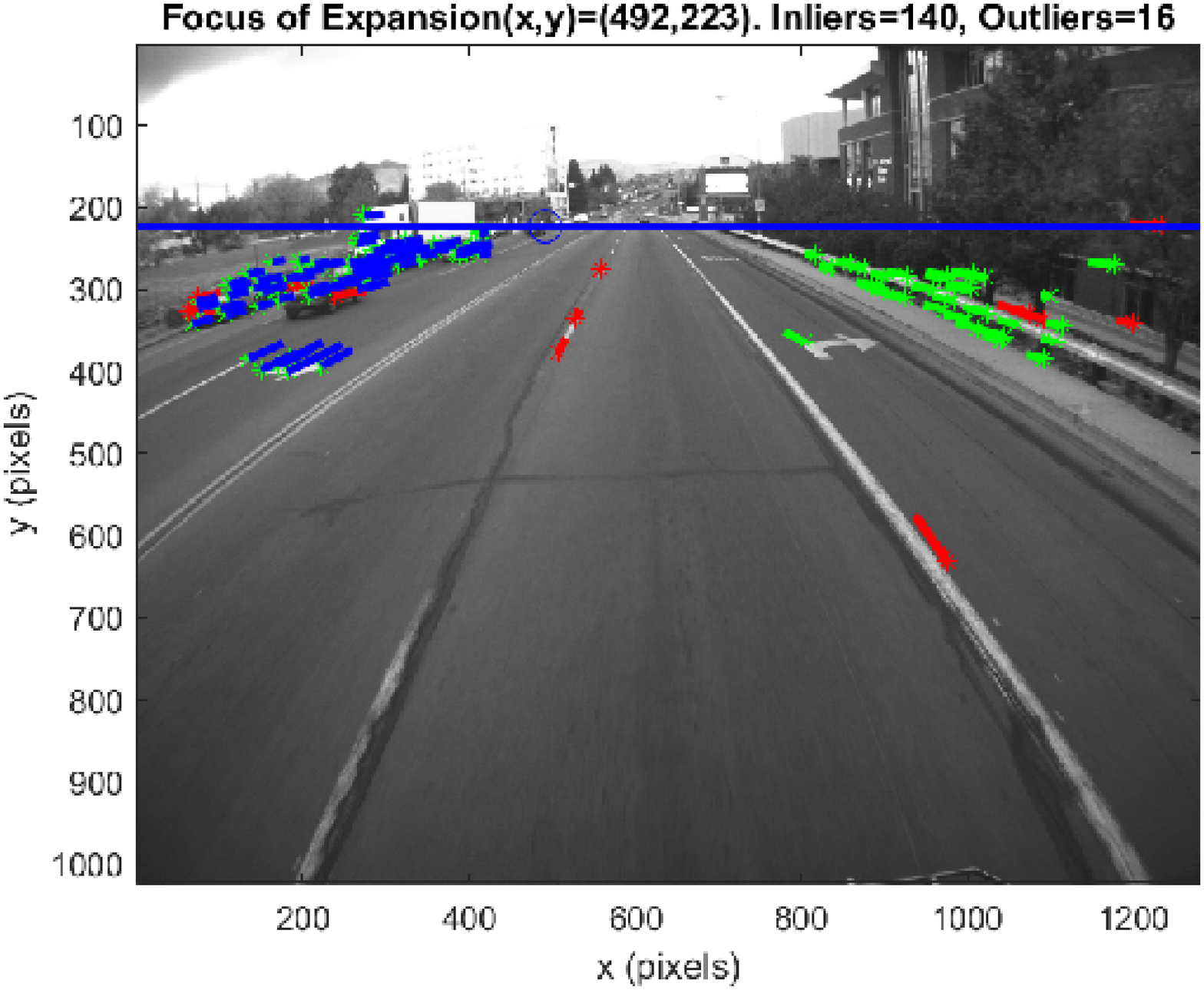}&\includegraphics[width=1.4 in, height=1 in]{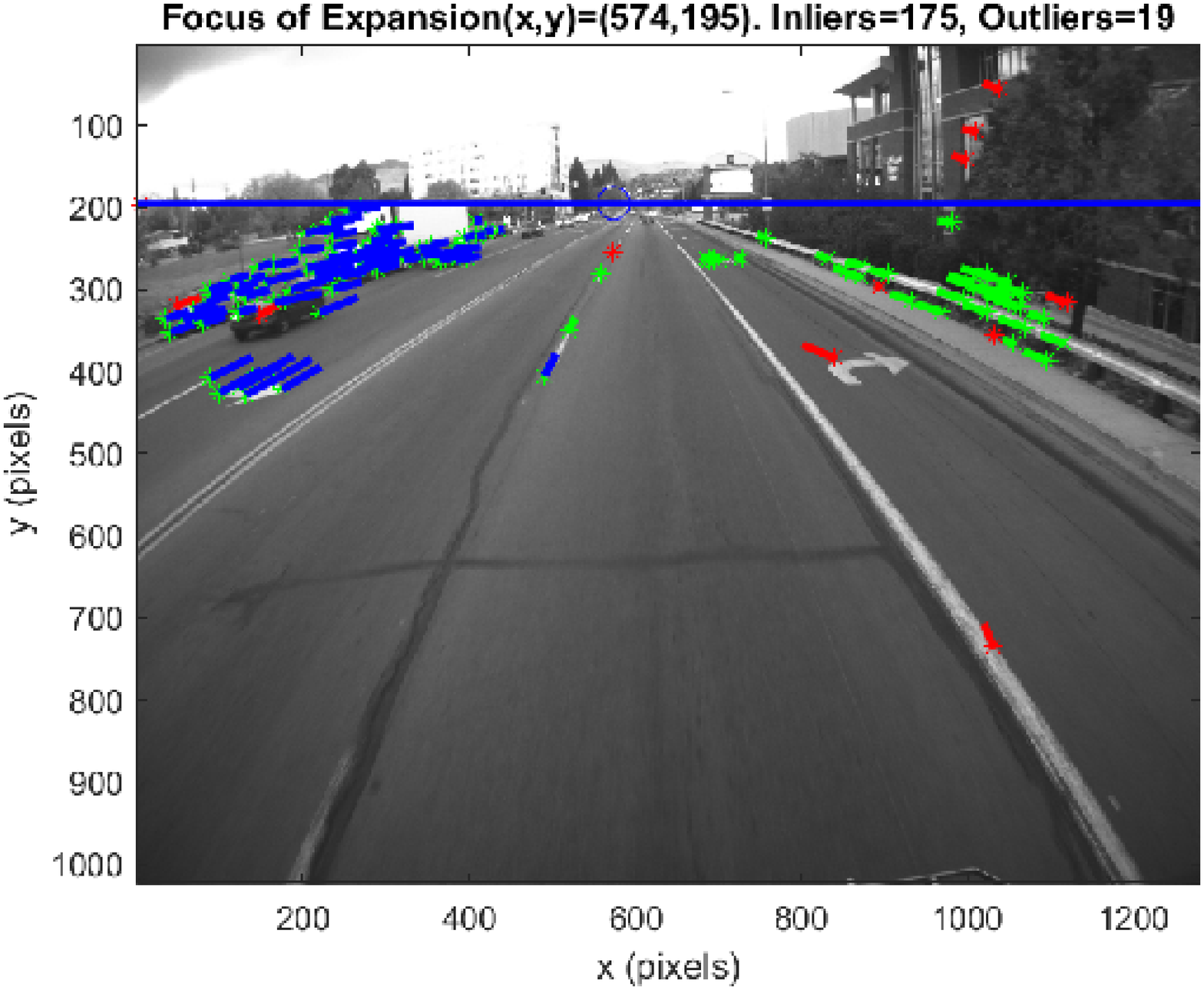}&\includegraphics[width=1.4 in, height=1 in]{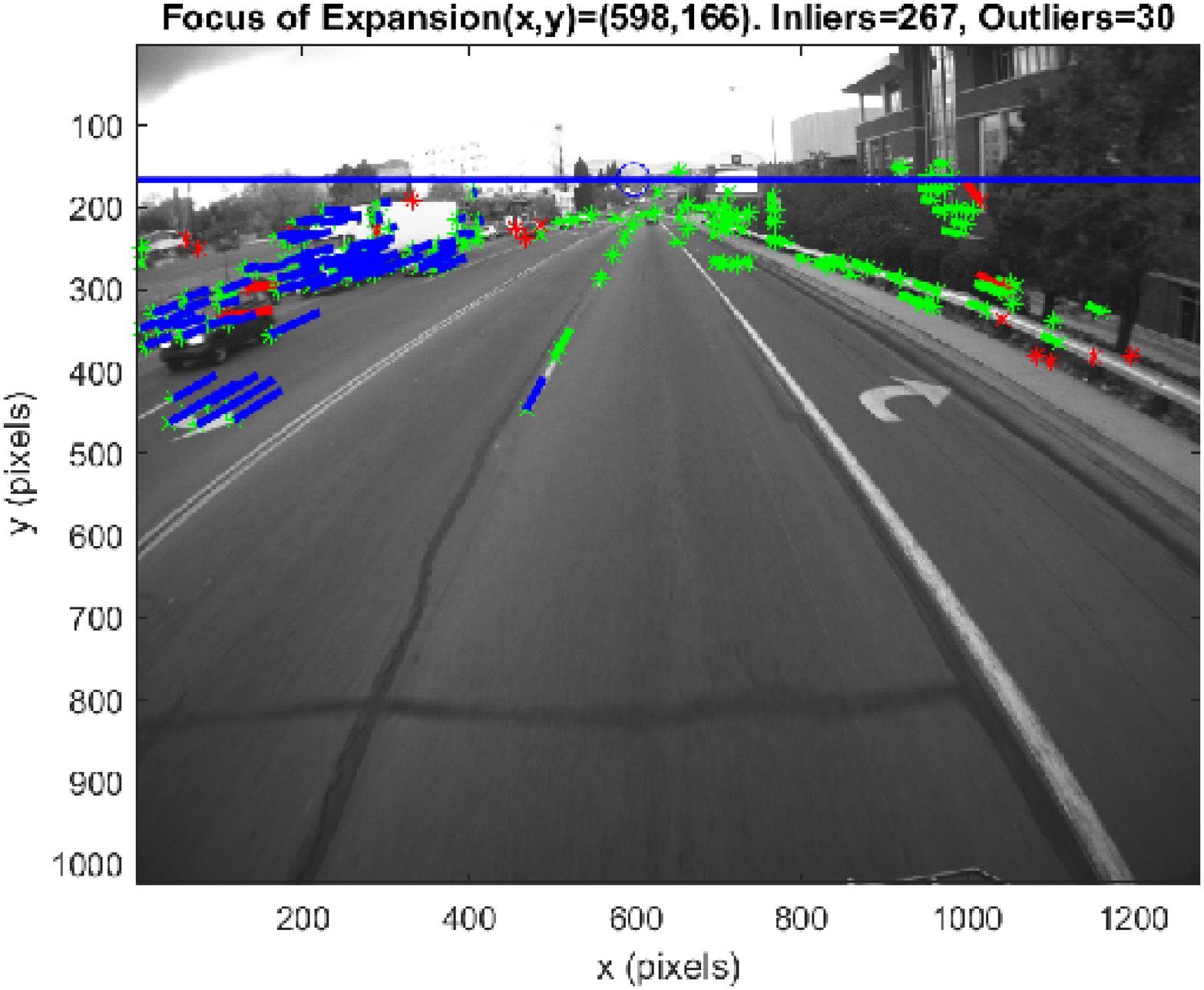}\\
{\small (e)}&{\small (f)}&{\small (g)}&{\small (h)}\\
\end{tabular}
\end{center}
\caption{(a) - (d) Raw images; (e) - (h) Results from the proposed background subtraction method. Red vectors denote outliers; Green vectors denote inliers; Blue vectors denote moving objects.}
	\label{results_two}
\end{figure*}

Figures \ref{results_one} and \ref{results_two} depict the results obtained from the moving camera onboard the electric bus. Figures \ref{results_one} (a) - (d) show the raw images and Figs. \ref{results_one} (e) - (h) show the estimated moving and non-moving features in the image plane. The same arrangement appears in Fig. \ref{results_two}. RGB color coding denotes outliers, inliers, and moving objects, respectively\footnote[1]{A video of the results obtained can be found in: https://goo.gl/HGgGUW}. Although the velocity of the bus is not constant and at the same time unknown, the results obtained are very accurate with the exception of some few individual false moving vector detections. 

For this research we used OpenCV and the C++ programming language to implement the LK optical flow algorithm, and the Machine Vision Toolbox \cite{Corke11} for modeling and simulating feature points and estimating the parameters of multiple linear regression. Modeling accurately feature points based on the FOE of the camera motion and camera intrinsics, and implementing multiple linear regression is critical in obtaining accurate results.

\section{CONCLUSIONS AND FUTURE WORK}

Background subtraction with a moving camera is a highly challenging problem that has gained popularity the last years due to the increasing interest in autonomous vehicles. In this paper, we addressed the problem using only visual information. In this research, we exploited the fixed camera height from the ground plane with the view to model and simulate feature points as seen by the camera sensor based on the intrinsic parameters obtained from the camera. A multiple linear regression model is applied to the synthetic data given the image coordinates of the features as well as the magnitude of optical flow vectors and is juxtaposed against the real data to identify significant optical flow variations that will signify moving object representations.

The developed methodology will be a core module that will be used to a) robustify GPS-denied simultaneous localization and mapping~\cite{VSEP_ICRA_2018,TUNNEL_AEROCONF_2018,NIR_ICUAS_2017,RHEM_ICRA_2017,LMAS_MSC_14} onboard the electric bus and other autonomous vehicles, and b) enable fast reactive planning in urban conditions with challenging traffic.  




\bibliographystyle{IEEEbib}
\bibliography{strings,paper}

\end{document}